\documentclass{edm_article}
\usepackage{hyperref}
\usepackage{microtype}

\begin{document}

\title{A Constraints-Based Approach to Fully Interpretable Neural Networks for Detecting Learner Behaviors}

\numberofauthors{2}


\author{
\alignauthor
Juan D. Pinto\\
       \affaddr{University of Illinois Urbana-Champaign}\\
       \email{jdpinto2@illinois.edu}
\alignauthor
Luc Paquette\\
       \affaddr{University of Illinois Urbana-Champaign}\\
       \email{lpaq@illinois.edu}
}

\maketitle

\begin{abstract}
    The increasing use of complex machine learning models in education has led to concerns about their interpretability, which in turn has spurred interest in developing explainability techniques that are both faithful to the model's inner workings and intelligible to human end-users. In this paper, we describe a novel approach to creating a neural-network-based behavior detection model that is interpretable by design. Our model is fully interpretable, meaning that the parameters we extract for our explanations have a clear interpretation, fully capture the model's learned knowledge about the learner behavior of interest, and can be used to create explanations that are both faithful and intelligible. We achieve this by implementing a series of constraints to the model that both simplify its inference process and bring it closer to a human conception of the task at hand. We train the model to detect gaming-the-system behavior, evaluate its performance on this task, and compare its learned patterns to those identified by human experts. Our results show that the model is successfully able to learn patterns indicative of gaming-the-system behavior while providing evidence for fully interpretable explanations. We discuss the implications of our approach and suggest ways to evaluate explainability using a human-grounded approach.
\end{abstract}


\section{Introduction}

The field of education, as with many other areas of research, has continued to see an increasing use of complex machine learning models as these have become more powerful and versatile over the years. However, as these models have grown in complexity, they have also become more difficult to interpret, leading to concerns around fairness, accountability, and trust \cite{khosraviexplainableartificialintelligence2022}, while also obscuring pedagogical insights that could improve learning outcomes among students.

There has been significant interest in tackling these issues within the eXplainable AI (XAI) community, with a growing group of educational data mining (EDM) researchers focusing on the implications and possible solutions to the problems arising from highly complex models (as evidenced by workshops specializing on XAI in education \cite{pintoPrefaceProceedingsHumanCentric2024}). There is growing awareness of the inherent limitations of post-hoc explainability techniques (i.e. generating explanations, typically feature importance measures, based on analyses of a model's inputs and outputs) which may often make them unsuitable for use in educational settings.

In this paper, we describe a novel approach to training a neural-network-based behavior detection model (more specifically, a convolutional neural network) that is interpretable by design \cite{swamyfuturehumancentricexplainable2023}. Our model is \emph{fully interpretable}, by which we mean that the parameters we extract for our explanations (1) have a clear interpretation, (2) fully capture the model's learned knowledge about the learner behavior of interest, and (3) can be used to create explanations that are both faithful and intelligible. We achieve this by implementing a series of constraints to the model that both simplify its inference process and bring it closer to a human conception of the task at hand.

We specifically focus on the detection of gaming-the-system behavior, a type of learner disengagement with an educational task. We chose this due to the existence of previous expert-based features and models \cite{paquetteunderstandingexpertcoding2014} that serve as a baseline of both accuracy and interpretability. We also chose to use a deep learning model due to their rising popularity among EDM researchers and the difficulty of faithfully interpreting their decision-making process \cite{pintoDeepLearningEducational2024b}.

Our research is guided by the following questions:
\begin{enumerate}
    \item Can we train a convolutional neural network to detect gaming-the-system behavior with convolutional filters acting as explicit behavioral patterns?
    \item Can we alter the model's architecture so that the patterns are the only learnable parameters?
    \item How can we create a differentiable perfect matching function that allows the model to definitively indicate a match or non-match based on the presence of any exact filter match to the input?
    \item How do the patterns learned by our model compare to those identified by human experts?
    \item How can we evaluate the interpretability of our model's explanations?
\end{enumerate}

\section{Background}

\subsection{Intrinsically interpretable models}

The literature abounds with descriptions of the differences between models that are intrinsically interpretable and those that are opaque (often referred to as ``black-box models'') \cite{liuApplicationsExplainableAI2024,kumarOverviewExplainableAI2023}. Yet fundamentally, any model's interpretability must be judged by the interpretability of the explanations derived from it \cite{rizzoTheoreticalFrameworkAI2023,pintoUnifiedFrameworkEvaluating2024}. These explanations can be derived from one of two sources: the model's inner workings themselves (e.g. its parameters or gradients) or via post-hoc techniques that use simplified approximations based on the model's inputs and outputs alone (such as LIME \cite{ribeirowhyshouldtrust2016} or SHAP \cite{lundbergunifiedapproachinterpreting2017}).

With this understanding, \cite{rizzoTheoreticalFrameworkAI2023} has proposed defining an explanation as the output of an interpretation function being performed on a piece of evidence. Evidence, in this context, refers to direct aspects of the model's workings, which can be extracted from any combination of its parameters, gradients, inputs, and outputs. A piece of evidence in itself carries no semantic significance---this can only be added by an interpretation function. The interpretation function describes how the model makes use of the evidence. By this definition, an explanation is therefore an inference made from interpreting specific evidence.

An important concept that \cite{rizzoTheoreticalFrameworkAI2023} introduce is that of \emph{explanatory potential}. This is the extent to which a specific set of evidences accounts for the whole of the model's predictions. In other words, using a subset of a model's parameters to create an explanation may be sufficient to explain a portion of the transformations that the input goes through to reach the output, say 70\%, but it may not provide a complete picture. Of course, a model does not typically use each parameter equally when making predictions, so the explanatory potential of each individual piece of evidence (along with their interactions) may vary.

It should be noted that explanatory potential as defined by \cite{rizzoTheoreticalFrameworkAI2023} pertains only to a set of evidences. It is the upper limit of the extent to which an explanation derived from it can account for the model's predictions, but an explanation also depends on an accurate interpretation function. Interpretations can be particularly challenging to identify given sufficiently complex transformations. Furthermore, even when derived from sound interpretations and a set of evidences with high explanatory potential, an explanation may still prove to be too complex for a target audience to understand. Thus why creating useful explanations must ultimately be a human-centered endeavor.

With these considerations in mind, the question then arises of how to evaluate not just an explanation's \emph{potential} explainability, but its actual usefulness in practice.

\subsection{Evaluating explainability}

Predictive and inferential models are typically evaluated primarily based on the accuracy of their outputs. Many robust methods for measuring different aspects of this accuracy have been devised and validated. It has become clear that no single accuracy metric captures the full complexity of a model's performance, and so a variety of metrics are used to provide a more complete picture \cite{boschMetricsDiscreteStudent2018,pelanekMetricsEvaluationStudent2015}. When it comes to evaluating explainability, however, there is yet no clear consensus.

This may be in part due to the subjective nature of explainability. While accuracy can be measured by comparing a model's predictions to a ground truth, explainability is a more abstract concept that depends on the needs and expectations of one or more end-users \cite{sureshExpertiseRolesFramework2021}. Still, some helpful frameworks have been proposed to guide the evaluation of explanations, both theoretical and practical.

\subsubsection{Theoretical framework}

\cite{pintoUnifiedFrameworkEvaluating2024} proposed a theoretical framework that brings together important elements to consider. They describe two main criteria through which explanations can be evaluated, with both as prerequisites for a \emph{useful} explanation: \emph{intelligibility} and \emph{faithfulness}. Intelligibility refers to the ease with which a human can understand the explanation (also referred to as \emph{comprehensibility} \cite{carvalhoMachineLearningInterpretability2019}), while faithfulness refers to the extent to which the explanation accurately reflects the model's inner workings. The authors argue that both criteria are necessary for a useful explanation, since an explanation that is not intelligible will likely not be used, and an explanation that is not faithful may be misleading.

As prerequisites to these two criteria, \cite{pintoUnifiedFrameworkEvaluating2024} further propose that an intelligible explanation must be \emph{plausible} (i.e. it aligns with human intuition \cite{jacoviFaithfullyInterpretableNLP2020}) while a faithful explanation must be \emph{stable} (i.e. it does not change drastically with small perturbations to the model's input \cite{alvarezmelisRobustInterpretabilitySelfexplaining2018}).

These criteria help explain why post-hoc explainability techniques, which are popular due to their ease of use and model-agnostic nature, are often problematic. Their faithfulness is difficult to measure with certainty since they rely exclusively on a model's inputs and outputs, with no direct access to its internal evidences, thus treating it precisely like a black box \cite{rudinstopexplainingblack2019}. Furthermore, while individual post-hoc techniques may lead to internally stable predictions, different approaches often produce different explanations for the same model and inputs \cite{krishnadisagreementproblemexplainable2022}. Their appeal is partly due to their ability to produce plausible and intelligible explanations---but when coupled with a lack of faithfulness, this may lead to problems where even experts are unable to decide which explanation to trust \cite{swamytrustingexplainersteacher2023}.

\subsubsection{Practical framework}

On the practical side of evaluation, \cite{doshi-velezrigorousscienceinterpretable2017} have proposed three methodological categories, as well as specific approaches, for evaluating explanations. Each evaluation approach emphasizes a different set of criteria \cite{pintoUnifiedFrameworkEvaluating2024}.

First, application-grounded evaluation involves testing a model in the real world for the target application for which it was developed. Approaches in this category can be costly and time consuming but have high fidelity with the needs of end-users. For example, learning dashboards are sometimes evaluated based on how well they help instructors understand their students' learning and provide help \cite{scheersInteractiveExplainableAdvising2021,tissenbaumSupportingClassroomOrchestration2019}. Such evaluation methods can effectively measure explanation intelligibility, but they do not directly measure faithfulness.

Second, human-grounded evaluation involves measuring how well humans can accurately answer questions about the model based on its explanation. Examples of the types of experiments in this category include: binary forced choice, where participants must pick which explanation they consider best when presented with multiple options (e.g. \cite{swamytrustingexplainersteacher2023}); forward simulation, where participants must correctly predict the model's output given specific inputs; and counterfactual simulation, where participants must correctly identify how a specific input needs to be changed in order to alter the model's given output. The latter two approaches in particular can serve as rigorous tests of the faithfulness of an explanation, while simultaneously capturing many aspects of intelligibility.

Finally, functionally grounded evaluation involves measuring some abstract aspects of a model or its explanation that capture constructs related to interpretability, but without human involvement. Being the least direct category, such approaches make it difficult to make robust claims about either intelligibility or faithfulness, but some measures such as model sparsity or explanation simplicity \cite{nguyenQuantitativeAspectsModel2020} can be said to make proxy measures of intelligibility. They can likewise be used to measure stability.

The modeling approach we describe in this paper will particularly emphasize faithfulness. That is, we are interested in training a model that makes it possible to generate explanations that fully capture the model's inference process, all while not becoming overwhelming and thus remaining intelligible. This is what we refer to as \emph{full interpretability}. For this reason, we also aim for explanations that can be evaluated using human-grounded methods, especially forward simulation and counterfactual simulation.

\subsection{Interpretable gaming behavior detection}

Gaming the system is a well-studied learner behavior in which learners exploit the properties of a learning environment in order to succeed at a task, often by guessing or extracting answers from a support system. Using sequences of five student-actions from Cognitive Tutor Algebra that were previously coded for gaming behavior by \cite{bakerlabelingstudentbehavior2008}, \cite{paquetteunderstandingexpertcoding2014} set out to create a classification model based on human expert insights.

Through cognitive task analysis---in which an expert in the behavior of interest reasons through various action sequences and explains their thinking out loud---\cite{paquetteunderstandingexpertcoding2014} were able to identify a set of features designed to capture elements of students' problem-solving behavior that experts pay attention to when looking for gaming-the-system behavior. For example, these features included whether a student reuses the same answer in a different part of the problem interface or if the student enters consecutive similar answers. Each feature is binary, indicating whether it was observed or not for a given action step. An action step is a point in time at which data was collected, and is triggered by the student either making a submission or asking for a hint.

With iterative input from the expert, they then used these features to create a set of 13 patterns indicative of gaming-the-system behavior. Using these patterns as sliding windows of consecutive actions in each sequence---thus serving as a cognitive model---the researchers achieved a Cohen's kappa of 0.430 on the training data and 0.330 on unseen test data. Their model was able to outperform a machine learning model trained on the same data by \cite{bakerlabelingstudentbehavior2008} that achieved a kappa of 0.218 and AUC of 0.691 on held-out test data (as reported in \cite{paquetteComparingMachineLearning2019}).

\begin{figure}
    \Description{Example expert pattern from cognitive model.}
    \centering
    \includegraphics[width=0.8\columnwidth]{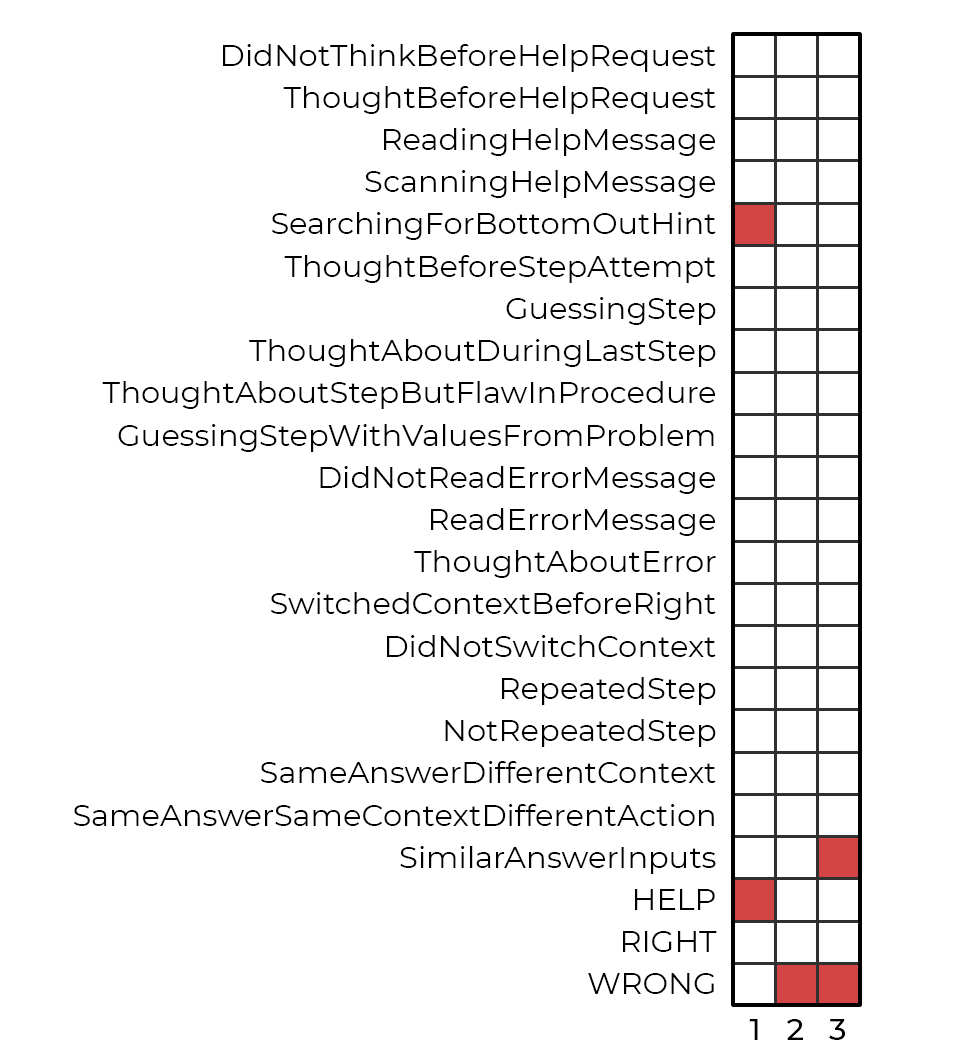}
    \caption{Example expert pattern from cognitive model.}
    \label{fig:expert_pattern}
\end{figure}

An example of one of these patterns can be seen in Figure~\ref{fig:expert_pattern}. Each row is a different feature identified via cognitive task analysis, and each column indicates a student action step. The pattern is a 3-action sequence. As originally written out, this pattern is as follows:

\begin{quote}
\itshape
\textbf{help} \& [searching for bottom-out hint] $ \rightarrow $ \textbf{incorrect} $ \rightarrow $ [similar answer] \& \textbf{incorrect}
\end{quote}

\cite{paquetteComparingMachineLearning2019} later trained a series of machine learning models using features that were automatically engineered using the results of the cognitive task analysis. Their best model, an ensemble of naive Bayes classifiers, achieved a kappa of 0.376 and AUC of 0.876 on the same test set. Importantly, they compared the interpretability of all three models trained and tested on the same data: the original ML model \cite{bakerlabelingstudentbehavior2008}, the cognitive model \cite{paquetteunderstandingexpertcoding2014}, and the hybrid approach \cite{paquetteComparingMachineLearning2019}. Not surprisingly, they found the cognitive model to be the most interpretable. The hybrid model that used expert-informed features provided an interesting middle ground, but the simple expert patterns that made up the cognitive model proved to be the most effective for human end-users. The ML model in this case was the least accurate and interpretable, but it also required far fewer resources to create.

\begin{figure*}[t]
    \Description{Model architecture.}
    \centering
    \includegraphics[width=\textwidth]{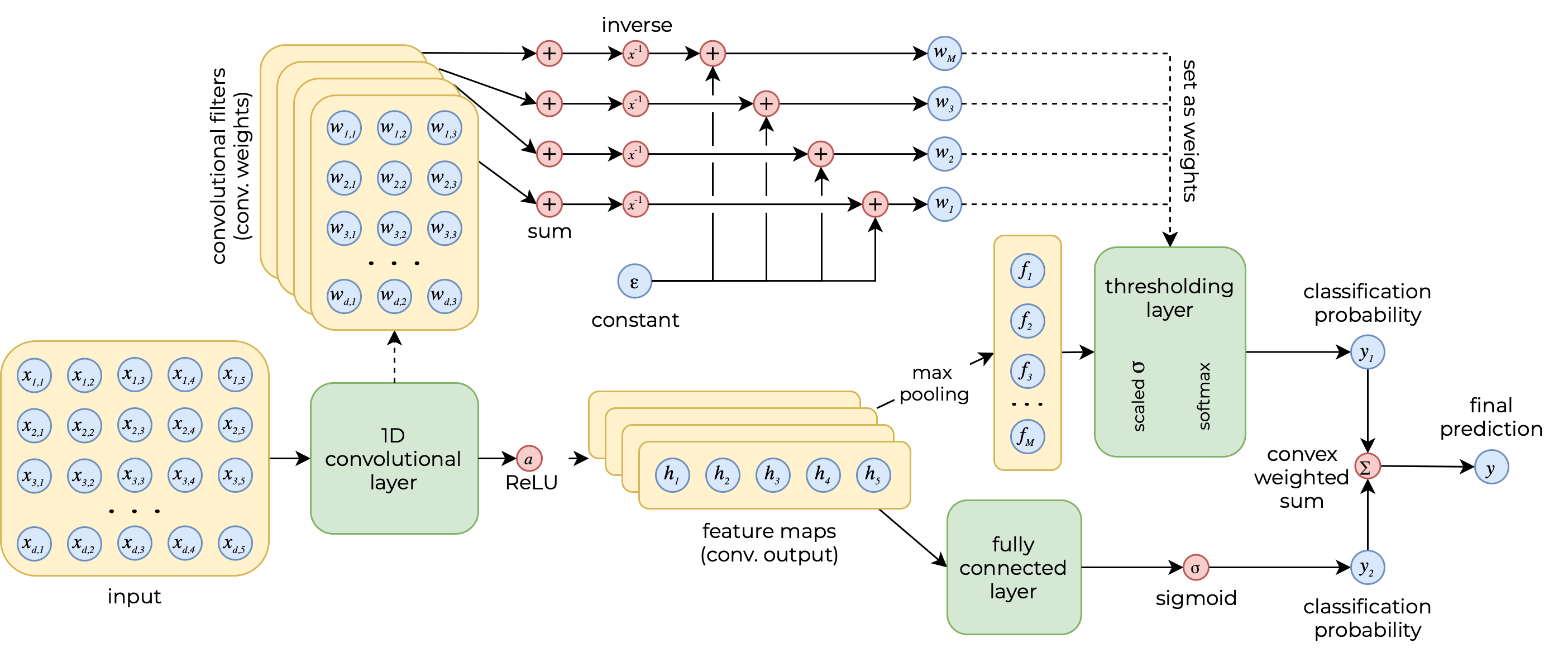}
    \caption{Model architecture.}
    \label{fig:architecture}
\end{figure*}

An analysis of the generalizability of the same 13 expert patterns explored how their frequency varied across different student populations and learning environments \cite{paquetteVariationsGamingBehaviors2017}. They found that differences in the learning environments were more strongly associated with differences in gaming behavior than were student populations. A similar study found that the cognitive model generalized better across learning environments than the ML model \cite{paquetteCrossSystemTransferMachine2015}. This suggests that the expert patterns identified by cognitive task analysis, while (not surprisingly) affected by aspects of the data from which they were generated, are nevertheless robust and can be used across contexts.

\section{Methods}

Rather than relying on post-hoc explainability techniques to create explanations with questionable faithfulness, we set out to create a neural-network-based behavior detection model that is interpretable by design \cite{swamyfuturehumancentricexplainable2023}. Furthermore, we aimed to create a model that is \emph{fully interpretable}. Such a model would be able to provide sufficient evidence to lend 100\% explanatory potential while ensuring that this evidence would have a clear interpretation. We deemed these qualities to be prerequisites to creating effective explanations that are both entirely faithful to the model's inner workings and intelligible to human end-users.

For this task, we use the same dataset and features as \cite{paquetteunderstandingexpertcoding2014} to train our model for two primary reasons. First, this allows us to directly compare the accuracy of our model with that of their cognitive model. Second, it gives us a template for the types of patterns we would like our model to be able to create. These patterns serve as the cognitive model's explanations---which are both fully faithful, since they make up the entirety of the cognitive model, and were manually designed by humans, suggesting a high level of intelligibility---so we reasoned that they would serve as a useful target for our own explanations. This also allows us to directly measure the similarities and differences between our model's learned patterns and those identified by human experts.

The dataset consists of sequences of actions from 59 students using the Cognitive Tutor Algebra system during an entire school year. Cognitive Tutor tasks students with solving multi-step mathematical problems and can provide on-demand hints. A total of 10,397 clips (i.e., student action sequences) were previously labeled by an expert \cite{bakerlabelingstudentbehavior2008} and contained 708 examples of gaming-the-system behavior (6.8\%).

Our model is purposefully simple, consisting of a single convolutional layer followed by a novel layer that we refer to as a \emph{differentiable thresholding fully connected layer} (explained in detail in section 3.2, see also Figure~\ref{fig:architecture}). The convolutional layer is designed to learn student patterns indicative of gaming-the-system behavior and which are similar in form to those identified by human experts in \cite{paquetteunderstandingexpertcoding2014}.

For the sake of simplicity, we only use student action sequences from the dataset that have a fixed length of 5 action steps (85\% of all sequences). We use the same held-out test set as \cite{paquetteunderstandingexpertcoding2014}, consisting of 25\% of the data. We further split the remaining data into a training/validation set using an 80/20 stratified split, using the validation set to tune hyperparameters and further refine our learned patterns. The final sets consist of 5,249 training clips, 1,313 validation clips, and 2,170 test clips, each with about 6\% positively labeled instances.

Since convolutional filters must have a fixed kernel length (i.e. number of action steps), we use a kernel length of 3, corresponding with the most common pattern length identified by the human expert. Theoretically, the model could learn patterns with smaller lengths than this, but not longer. We set the convolutional layer's padding to 1 to allow the model to learn shorter patterns on the edges of input action sequences, and we set the number of filters to 2,048 to allow the model to learn a wide variety of patterns.

During training, the model contains additional elements, such as a dropout layer for the outputs of the convolutional layer, as well as a fully connected layer that acts as a traditional classifier via a weighted branching architecture, but these are not used during inference.

To achieve full interpretability, we followed two atypical approaches. First, our model architecture is minimalist---i.e. it avoids any learnable weights outside of the target patterns contained in the convolutional layer's filters, such as bias terms or fully-connected-layer weights. Second, we introduced a series of constraints to the model's learning process that encourage the convolutional weights to follow certain guidelines. Both of these approaches allow us to control the amount of inherent flexibility in the model.

Flexibility is key to models that aim to be interpretable by design. Closely related to the tradeoff between bias and variance, increased flexibility (which is itself related to decreased bias) tends to lead to more complex models that make interpretability more difficult. Thus, our constraints are designed to decrease flexibility just enough to allow the model to learn the patterns we are interested in, while still maintaining a high level of predictive accuracy.

\subsection{Loss function constraints}

The model was trained using binary cross entropy as its main loss term. To this were added four regularization terms, each aimed to constrain the model's learning in a specific way. The full loss function is given by:
\[
    L = L_{\text{BCE}} + \gamma_{1} L_{\text{bin}} + \gamma_{2} L_{\text{min}} + \gamma_{3} L_{\text{sub}} + \gamma_{4} L_{\text{poss}}
\]
where $L_{BCE}$ is the main loss term (binary cross entropy loss) and $\gamma_{i}$ are scaling weights that control the impact of each regularization term.

All of these constraints specifically affect the weights in the model's convolutional layer. Figure~\ref{fig:constraints} visually demonstrates the impact of each additional constraint on the convolutional filters. We will now describe each of these regularization terms in turn.

\subsubsection{Binarize convolutional filter weights}

\begin{figure}
    \Description{Visualization of the effects of increasing constraints on the model's convolutional filters.}
    \centering
    \includegraphics[width=\columnwidth]{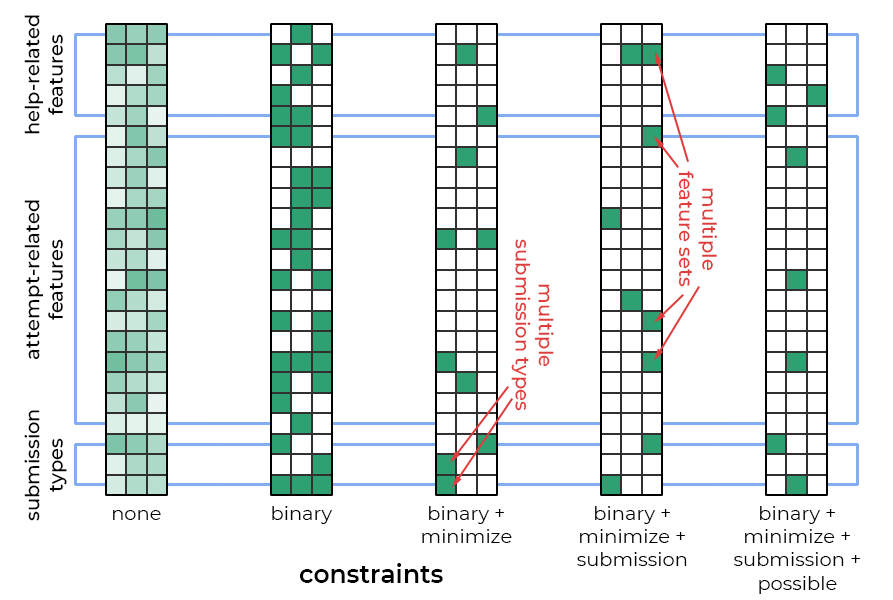}
    \caption{Visualization of the effects of increasing constraints on the model's convolutional filters.}
    \label{fig:constraints}
\end{figure}

The first regularization term constrains the model's convolutional layer to learn weights very close to $0$ or $1$ by penalizing values that deviate from these. Constraining the weights in this way enables a more straight-forward interpretation of each filter as a sequential pattern that emulates the binary presence or absence of specific features per action step. This binarizing effect can be seen in the difference between the two left-most filters in Figure~\ref{fig:constraints}.

For this constraint, we used the term described in \cite{jiangpredictivesequentialpattern2021} for multiple concurrent elements:
\[
    L_{\text{bin}} = \sum_{p=1}^M \sum_{n=1}^k \sum_{j=1}^d \left| W_{pnj}^2 - W_{pnj} \right|
\]
where $ W \in \mathbb{R}^{M \times k \times d} $ is the weight tensor for the convolution layer, $ M $ is the number of filters (i.e. the number of patterns to learn), $ k $ is the number of action steps, and $ d $ is the number of features.

Note that this approach is not simply a hard thresholding of the weights (such as rounding) since that would be a non-differentiable operation---i.e. it would lack a well-defined derivative, creating complications for the back-propagation process used to train the model. While the absolute value function in our regularization term is also non-differentiable at exactly $ 0 $, it remains differentiable for every other case, allowing the model to learn via gradient descent.

\subsubsection{Minimize positive weights per action step}

An additional regularization term places a penalty on action steps that have a high number of positive weights. We introduced this term to encourage the model to focus only on the most relevant features per action step. We reasoned that having fewer activated features would also improve intelligibility by reducing cognitive load at explanation time.

The impact of this constraint can be seen in Figure~\ref{fig:constraints}: there are far fewer positive features in the third filter from the left compared to the second. Note that the weights in both of these example filters are already binarized.

To achieve this, we added the following term to the loss function:
\[
    L_{\text{min}} = \sum_{p=1}^M \sum_{n=1}^k \text{ReLU}(r^{(a - w_{pn})} - b)
\]
where $ r $ controls the penalty rate, $ a $ sets the number of activated features at which the penalty begins to be applied, and $ b $ is a bias that further accelerates the penalty rate. These hyperparameters were set during tuning.

\subsubsection{Force single submission type}

In our dataset, every student action step has one submission type attached, whether a request for \emph{help} in the form of a next-step hint, a \emph{correct attempt}, or an \emph{incorrect attempt}. Consequently, we introduced a regularization term to ensure that the model learns at most one submission type per action step.

Note that this is slightly simplified from the original features described in \cite{paquetteunderstandingexpertcoding2014}, which used two separate \emph{incorrect attempt} labels: \emph{wrong} and \emph{bug}. Some of the expert patterns also use \emph{attempt} to indicate any submission type that is not \emph{help}. We removed this feature to simplify the submission types---all expert patterns that used it also included a non-help-related feature in the same action step, making it unnecessary.

The regularization term we used to enforce this constraint is:
\[
    L_{\text{sub}} = \text{ReLU}(s_{pn} - 1)
\]
where $s_{pn}$ refers to a squashed vector of a slice of $W_{pnj}$ across dimension $j$---similar to the penalty term to minimize the number of positive weights as described above, but using only the features that correspond to submission types.

Figure~\ref{fig:constraints} again shows the impact of this regularization term. The fourth filter from the left (where this constraint has been introduced) contains at most one positive submission type per action step (column), unlike those to its left which don't have this constraint.

\subsubsection{Ensure possible feature combinations}

Our final regularization term encourages the model to learn only positive weights in one of two feature sets for any single action step. For example, the nature of our feature set makes it so that the first five features are help-related features, so they can only be positive when the submission type is \emph{help} (third feature from the bottom in the filters visualized in Figure~\ref{fig:constraints}). By extension, all other features (except submission types) are attempt-related features and can only be activated when the submission type is NOT \emph{help}.

This means that the input data will never have a positive feature in both feature sets in the same action step. Filters that don't follow this guideline will never lead to a perfect match, so our constraint is designed to avoid such ineffective patterns. Compare the right-most filter in Figure~\ref{fig:constraints} with the others to see the effect of this constraint.

The regularization term we used for this purpose is:
\[
    L_{\text{poss}} = \sum_{p=1}^M \sum_{n=1}^k \min \left( \left(
        \frac{ \sum_{i \in {S_h}} W_{pni} }{ \text{len}(S_h) }
    \right)^2 , \left(
        \frac{ \sum_{i \in {S_a}} W_{pni} }{ \text{len}(S_a) }
    \right)^2 \right)
\]
where $ S_h $ and $ S_a $ are vectors containing the indices of help- and attempt-related features, respectively.

\subsection{Explicit filter matching}

Aside from the constraints implemented via regularization terms to the loss function, we introduced a set of final constraints on our model by altering the architecture of the model itself. In a typical convolutional neural network, the outputs of the convolutional layer, after passing through an activation function, become the inputs to a fully connected layer designed to turn them into a final probability prediction reminiscent of a logistic regression. As with the convolutional layer's weights, the weights of this fully connected layer are learned during training. Because these weights play an important role in shaping the model's final prediction, they would need to be included in the evidence used to create a fully faithful explanation.

In order to achieve our goal of allowing the convolutional filters to act as self-contained patterns of gaming-the-system behavior, and to ensure that a filter match unequivocally leads to a positive prediction and that the absence of a matching pattern leads to a negative prediction, we designed a novel fully connected layer architecture. Our layer diverges from the conventional linear transformation followed by a global activation function. Instead of this conventional approach, our layer employs an element-wise activation function---specifically, a scaled sigmoid that approximates a step function---applied immediately after the input is multiplied by the layer's weights and prior to the summation step.

The layer's inputs are the convolutional layer's outputs after passing through a max pooling layer that ensures only each filter's convolution with the most matching activations is passed on. The layer's weights (of length equal to the number of convolutional filters, $ M $, and thus equal to the length of the layer's input filter), rather than being learned, are manually set to the inverse of the sum of all weights for each filter.

In essence, this makes it so that when the layer's inputs are multiplied element-wise by its weights, any filters that perfectly matched on the input will result in a $ 1 $, whereas all others will result in a $ 0 $. Note that a ``perfect match'' in this case exists when all positive weights in a filter (i.e. $ 1 $ after binarization) are also positive in the input instance. Negative weights in a filter (i.e. $ 0 $ after binarization) can have any value in the input instance.

This design allows the network to output a scalar value that decisively indicates a \emph{match} (i.e., above or equal to 0.5) or \emph{non-match} (i.e., below 0.5) based on the presence of any exact filter match. We refer to this mechanism as a \emph{differentiable thresholding fully connected layer}, as it bridges the gap between hard thresholding (e.g. a conditional function) and standard fully connected layer soft aggregation, all while remaining trainable via gradient descent.

\subsubsection{Formal description}

Formally, we can describe the differentiable thresholding fully connected layer as follows.

Let the matrix $ h \in \mathbb{R}^{M \times C} $ be the output of the convolutional layer (the feature maps) after passing through a $ \text{ReLU} $ activation function, where $ M $ is the number of filters in the convolutional layer and $ C $ is the number of convolutions. A max pooling function is applied to these feature maps to keep only the convolution with the most activations per filter. Therefore, let $ f = \text{MaxPool}(h) $, where $ f \in \mathbb{R}^M $.

This vector then serves as the input to the differentiable thresholding fully connected layer. Let the layer's weight vector be $ w \in \mathbb{R}^M $. The thresholding layer processes the input in four distinct stages:

First, the weights $ w_p $ are manually set using the convolutional filters. If $ W \in \mathbb{R}^{M \times k \times d} $ is the weight tensor for the convolutional layer, then for each filter $ p $:
\[
w_p = \frac{1}{\sum_{n=1}^{k} \sum_{j=1}^{d} \bigl| W_{p,n,j} \bigr| + \varepsilon}
\]
where $ \varepsilon $ is a small constant added for numerical stability.

Second, each input element is multiplied by its corresponding weight:
\[
z_i = f_i \cdot w_i, \quad \text{for } i = 1, \ldots, M
\]

Third, a scaled sigmoid function is applied to $ z_i $ to approximate a step function:
\[
a_i = \sigma\Bigl(t (z_i-\beta)\Bigr)
\]
where $ \sigma(\cdot) $ is the sigmoid function, $ t $ controls the steepness (with larger $ t $ making $ \sigma $ approach a step function), and $ \beta $ is an offset (typically set near 1, e.g. 0.99).

Fourth, a softmax function with a small temperature $\tau$ is used to compute weights that emphasize larger activations:
\[
s_i = \frac{\exp\left(\frac{a_i}{\tau}\right)}{\sum_{j=1}^n \exp\left(\frac{a_j}{\tau}\right)}
\]

Finally, the layer's final output is a weighted sum of the activated values:
\[
y_{\text{thresholding}} = \sum_{i=1}^M a_i\, s_i
\]

This formulation allows the layer to perform a nearly hard thresholding operation in a differentiable manner. The sigmoid approximates a step function, ensuring the network can decide between \emph{match} (output $ \geq 0.5 $) and \emph{non-match} (output $ < 0.5 $) in a soft, gradient-friendly fashion.

\subsubsection{Branching architecture}

Notably, our architecture includes both our custom thresholding layer and a conventional fully connected layer, with the final output being a weighted sum of the two via a convex combination using the weight $ \alpha $. Like the $ \gamma_{i} $ scaling weights we use to modulate the regularization terms in our loss function, this branching approach allows us to control the impact of the thresholding layer. In this way, the model can have a more traditional warm-up period during training before our explicit filter matching via differentiable thresholding fully connected layer begins to influence the model's final predictions. Once this weight surpasses a certain (low) threshold, we also freeze the weights of the traditional fully connected layer so that the convolutional weights are the only parameters learned.

The traditional fully connected layer, or traditional branch of our architecture, processes the input through adaptive max pooling, flattening, a linear transformation, and a sigmoid activation:
\[
y_{\text{trad}} = \sigma\Bigl( \mathbf{w}^{\top} \, \text{flatten}\Bigl( \text{pool}(x) \Bigr) \Bigr)
\]

The model's final output is then computed as a weighted sum of the two branches:
\[
y = (1-\alpha) \, y_{\text{trad}} + \alpha \, y_{\text{thresholding}},
\]
with an additional clipping to ensure that $ y = \min \{ y, \, 1 \} $.

This approach serves as a differentiable mechanism that gradually shifts from a traditional fully connected layer to our custom thresholding layer as controlled by $ \alpha $.

\subsection{Progressive constraining}

In order to allow the model to properly learn over multiple epochs, we used the scaling weights $ \gamma_{i} $ to control the impact of each regularization term $ i $ independently, as well as $ \alpha $ to control the impact of the differentiable thresholding fully connected layer. We allowed a warm-up period for each constraint, during which time its corresponding weight was gradually increased from $ 0 $ to its final value.

The warm-up periods were staggered, with each term having its own starting epoch and growth rate, which we treated as hyperparameters to be tuned. We used this staggered approach so that the model could begin incorporating a single constraint at a time, having greater flexibility at the beginning of the training process and progressively becoming more constrained, until finally converging on our desired filter format and inference approach.

We used grid search to find the optimal values for these hyperparameters, ensuring both that the model reached appropriate accuracy on the validation set and that the convolutional filters followed the constraints we set out to enforce. The grid search was set up with various possible starting epochs for each constraint, different growth rates, and different orders in which to introduce the constraints. Alongside accuracy on the validation set, we also exported visualizations of the learned filters at different points during training to compare.

We found that when the binarizing constraint was introduced too early, the weights had difficulty changing, even when the loss was high due to the penalties from other constraints. We also found that when the explicit filter matching constraint (via the differentiable thresholding fully connected layer) was introduced too late, the model often struggled to later learn patterns that matched the inputs exactly. Ultimately, we found that the best results were achieved when we introduced the constraints in the following order: explicit filter matching, possible feature combinations, single submission type, minimum positive weights per action step, and binary filters.

\subsection{Training across multiple eras}

Unfortunately, even with the gradual introduction of constraints, the model eventually reached a point where the gradients stopped flowing. This was likely due to the combination of the binary constraint and the explicit filter matching---along with its scaled sigmoid approximating a step function---which prevented the model from continuing to learn to the point of overfitting.

To address this potential underfitting, we introduced a mechanism to reset the training process at regular intervals. Extrapolating from the conventional use of the term \emph{epoch}, borrowed from geologic chronology, we refer to these completed intervals of training epochs as training \emph{eras}.

\begin{figure}
    \Description{Scaling weight increases with staggered warm-up periods across two eras of 200 epochs each.}
    \centering
    \includegraphics[width=\columnwidth]{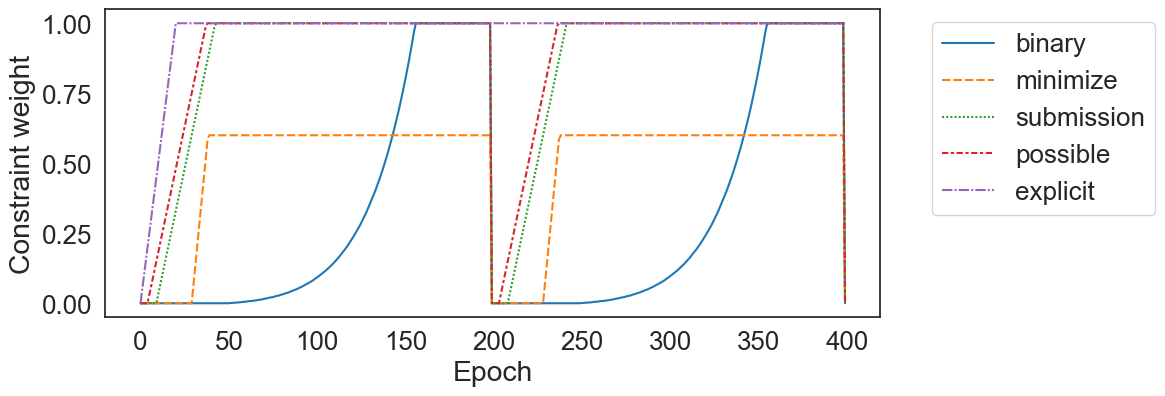}
    \caption{Scaling weight increases with staggered warm-up periods across two eras of 200 epochs each.}
    \label{fig:eras}
\end{figure}

We trained our model for a total of 50 eras, each consisting of 200 epochs. At the end of each era, we reinitialized the weights for empty filters (those with all $ 0 $ weights) and for those with precision < 0.3. We also reset all $ \gamma_{i} $ values, restarting their staggered warm-up periods. This approach allowed the model to learn entirely new gaming-the-system patterns or refine existing ones by once again allowing gradients to flow and giving the chance for the weights to escape local minima, all while still adhering to the constraints we set out to enforce. We did not reset the $ \alpha $ scaling weight. Figure~\ref{fig:eras} shows the model's progressive constraining across two eras.

One advantage of our minimalist architecture---one with learnable weights only in the convolutional layer---is that we could save these weights to an external file at the end of each era, retaining the model's entire set of learned parameters for future use. This provided us a far larger set of gaming-the-system patterns than what the model could learn in a single era. By analyzing these patterns, we were able to identify the most effective ones, remove duplicates, and insert those remaining back into the model for final inference. We further describe this process and report our results in the next section.

\section{Results}

Our approach generated a total of 102,400 filters (2,048 per era for 50 eras) from which we saved the 25,981 filters that achieved a precision above 0.3. Because the non-reinitialized filters were carried over to the initial state in subsequent eras, there was a large amount of repetition in the total set of filters identified during training. After removing duplicate filters, only 1,359 were unique. Furthermore, because of the nature of convolutional filters, if the positive weights in one filter are also positive in another, even if the latter has more positive weights, the former will match on the same inputs as the latter. In this scenario, the first filter has captured a more general gaming-the-system pattern than the second.

After accounting for these redundant filters, and keeping only the more general ones, we were left with 210 unique, non-redundant patterns.

\begin{figure}[h]
    \Description{Cumulative metrics on patterns sorted descending by precision on the training set.}
    \centering
    \includegraphics[width=\columnwidth]{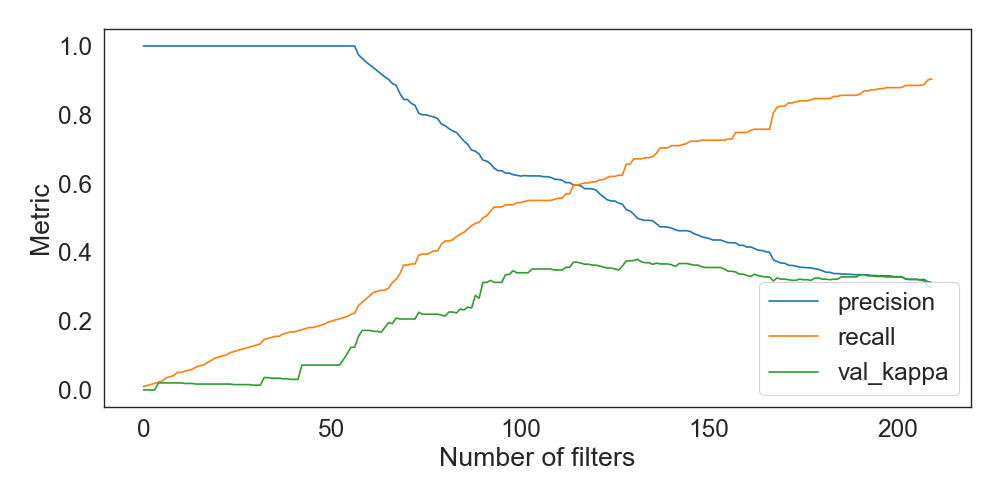}
    \caption{Cumulative metrics on patterns sorted descending by precision on the training set.}
    \label{fig:ablation}
\end{figure}

From this set of patterns, we individually calculated each one's precision on our training set, which ranged from 0.302 to 1.0 with a mean of 0.610. We sorted the patterns by precision and calculated the cumulative Cohen's kappa on our validation set for each subset of $ n $ best patterns. The resulting metrics are shown in Figure~\ref{fig:ablation}.

\begin{table}[ht]
    \centering
    \caption{Performance metrics on various datasets.}
    \begin{tabular}{lccccc}
    \hline
    Set   & Accuracy & AUC   & Kappa & Precision & Recall \\
    \hline
    train & 0.940    & 0.923 & 0.541 & 0.499     & 0.672  \\
    val   & 0.917    & 0.883 & 0.380 & 0.360     & 0.513  \\
    test  & 0.909    & 0.847 & 0.319 & 0.324     & 0.422  \\
    \hline
    \end{tabular}
\end{table}

Based on this analysis, we selected the top 132 patterns to use in our final model. We then evaluated the model's performance on our held-out test set, achieving a kappa of 0.319. This is slightly lower than the cognitive model's kappa of 0.330 on the same test set \cite{paquetteunderstandingexpertcoding2014}.

\subsection{Comparison with expert patterns}

We compared our model's 132 final learned filters with the expert patterns identified by \cite{paquetteunderstandingexpertcoding2014}. However, of these 13 expert patterns, we had to make some small modifications to two of them to make them fully compatible (and comparable) with our filters. The two expert patterns in question contained an additional condition not reflected directly in the binary features: ``\emph{with at least one similar answer between steps}''.

For example, one of these patterns was originally formulated as follows:
\begin{quote}
    \itshape
    \textbf{help} $ \rightarrow $ \textbf{incorrect} $ \rightarrow $ \textbf{incorrect} $ \rightarrow $ \textbf{incorrect} (with at least one similar answer between steps)
\end{quote}
which we split into two variations that include the feature \emph{similar answer} in key locations.:
\begin{quote}
    \itshape
    \textbf{help} $ \rightarrow $ \textbf{incorrect} $ \rightarrow $ \textbf{incorrect} \& [similar answer] $ \rightarrow $ \textbf{incorrect} \\ \\
    \textbf{help} $ \rightarrow $ \textbf{incorrect} $ \rightarrow $ \textbf{incorrect} $ \rightarrow $ \textbf{incorrect} \& [similar answer]
\end{quote}
This increased the number of expert patterns to compare up to 15.

Furthermore, because some expert patterns contained 2 or 4 action steps (as opposed to our filters' 3), we expanded them to all possible 3-action sequences for this analysis. From 2-action patterns we created two separate patterns: one with a blank action step at the beginning and another with a blank action step at the end. From 4-action patterns we also created two separate patterns: one encompassing the first three action steps and another the last three. While this latter scenario likely leads to trimmed patterns that the expert may no longer consider indicative of gaming-the-system behavior, we reasoned that such a comparison with our model's learned filters would still be informative.

\begin{figure}
    \Description{Expert patterns (red) and our model's learned  filters (green) for the most similar pairs.}
    \centering
    \includegraphics[width=\columnwidth]{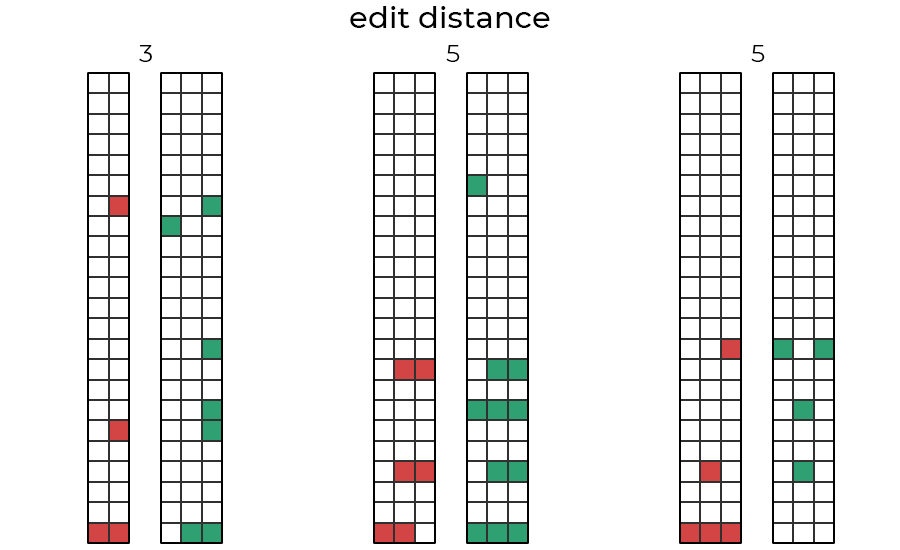}
    \caption{Expert patterns (red) and our model's learned  filters (green) for the most similar pairs.}
    \label{fig:similar_patterns}
\end{figure}

We made our comparison using the Levenshtein edit distance, which measures the minimum number of single-value edits (in our case, substitutions from $0$ to $1$ or vice versa) required to change one pattern into another. We found that the average distance between our filters and the expert patterns was 13.5, with a standard deviation of 2.6. This suggests that our model's learned filters differed quite significantly from those identified by human experts. Our model did not learn any of the expert patterns on its own. The most similar patterns had an edit distance of 3 and 5, but only three pairs were this similar (shown in Figure~\ref{fig:similar_patterns}).

One noticeable difference is that our model's learned filters tended to have many more positive features than the expert patterns.
The model's filters had a mean of 12.0 positive features, with a range of 4 to 18 and a standard deviation of 2.3. This is in contrast to the expert patterns, which---when taking into account only those with three action steps---had a mean of 4.9 positive features, with a range of 4 to 7 and a standard deviation of 1.1. The entire set of expert patterns had a mean of 5.1 positive features, with a range of 4 to 7 and a standard deviation of 1.0.

We also measured the differences between the model's own learned filters pairwise. The mean edit distance between two filters was 12.9, with a range of 2 to 26 and a standard deviation of 3.7. The smallest edit distance between two filters was 2, which occurred 13 times, while the next smallest edit distance was 3, which occurred 25 times.

\section{Discussion}

Our model achieved performance comparable with the cognitive model described in \cite{paquetteunderstandingexpertcoding2014}, with a slightly lower kappa on the held-out test set. Its precision was higher than the cognitive model's (0.324 vs. 0.307), while its recall was lower (0.422 vs. 0.528). It outperformed the machine learning model created by \cite{bakerlabelingstudentbehavior2008}. This suggests that our model was able to learn patterns indicative of gaming-the-system behavior.

Surprisingly, despite the many eras during which our model was trained and reinitialized with new starting weights, and despite the high number of convolutional filters (2,048), the final number of usable filters was only 210. Only 5.2\% of our total number of filters with precision > 0.3 were unique. The model therefore repeatedly learned many of the same filters, even while it had the opportunity to learn many new ones.

One possibility is that the constraints we introduced were too restrictive, causing the model to converge repeatedly on local minima. This is supported by the fact that the model's filters had more positive features than the expert patterns on average---the very outcome that our $ L_{\text{min}} $ regularization term tried to prevent.

Our more complex filters (i.e. with more positive features) help explain the model's higher precision but lower recall compared to the cognitive model. The model's filters were more specific, leading to fewer false positives, but they were also less encompassing, leading to more false negatives. This also explains the two models' similar overall performance despite our model having many times more patterns than the cognitive model.

Despite this, the final set of learned filters do successfully predict gaming-the-system behavior when compared to the cognitive model with similarly structured patterns. While our model uses many more patterns than the cognitive model, its performance on the held-out test set indicates that it did not overfit to the training data. Given this, the question remains of how well these filters can be used to create effective explanations.

\subsection{Explainability}

We set out to create a model that was fully interpretable, meaning that we could extract sufficient evidence from it to create an explanation that (1) is faithful, fully capturing the model's inference process, while (2) remaining intelligible.

By ensuring that the model's only learnable parameters are contained in the convolutional filters, and by constraining those filters to follow the template of patterns created by human experts, we believe we have achieved our goals. The weights of the convolutional layer provide 100\% explanatory potential \cite{rizzoTheoreticalFrameworkAI2023}. They can be clearly interpreted as sequential patterns that emulate the binary presence or absence of specific student actions.

\begin{figure}[h]
    \Description{Example explanations created from the model's learned filters.}
    \centering
    \includegraphics[width=\columnwidth]{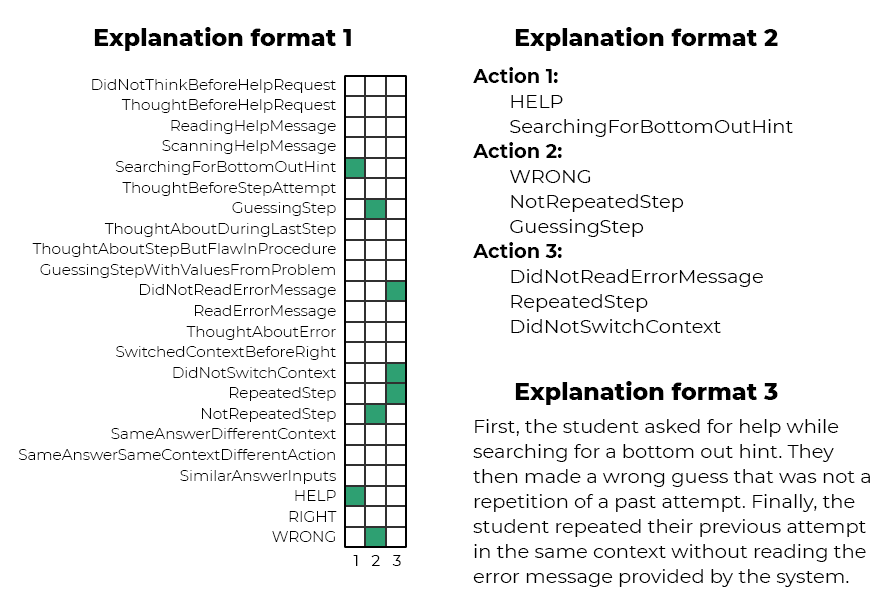}
    \caption{Example explanations created from the model's learned filters.}
    \label{fig:explanations}
\end{figure}

These weights, paired with this interpretation, can thus be used to create different kinds of explanations. For example, if the model detects gaming-the-system behavior in a student's actions because they matched a specific filter, that filter's weights can be used to explain why the model made that decision. These explanations can take many forms, such as visual matrices for more technical end-users, simplified bullet-point explanations, or even text-based explanations in user-friendly language using LLMs (see the examples in Figure~\ref{fig:explanations}).

This is an example of local explainability, where a specific prediction is explained. More global patterns can also be extracted from the model's filters, such as the most common patterns or the most important features. These can be used to create more general explanations, such as identifying the most common reasons a student is flagged for gaming-the-system behavior or the most important features to look for. Through all of this, the model's filters provide a direct link to the entirety of the model's learned parameters, allowing for fully faithful explanations.

The implications of this are significant. If such fully interpretable models are possible and can retain sufficient flexibility to fit a wide range of use cases, they could be used to create more transparent AI systems in education. Our contribution is not the specific model we have trained, but rather the general constraints-based approach we have outlined, along with the evidence we have provided that it functions as intended.

However, before making strong claims about interpretability, it is clear that we need to evaluate the explanations our approach can generate. Fortunately, the explanations we have proposed here are the perfect test case for the human-grounded evaluation methods outlined by \cite{doshi-velezrigorousscienceinterpretable2017}. There is also room for future work to further refine our approach.

\subsection{Limitations and future work}

While our approach so far indicates the successful learning of interpretable patterns for behavior detection, there remain important limitations to address. The most pressing of these is the issue we encountered with the model's gradients no longer flowing after a certain number of training epochs. Experiments we conducted with disabling individual constraints while tracking gradient norms indicated that the issue arises from the combination of the binary constraint and the explicit filter matching via thresholding layer.

We managed to produce a workable model with functioning filters through the use of multiple training eras, but this is not an ideal solution. There may be specific transformations where the gradients are being blocked that could easily be addressed. For example, it may simply be that the thresholding layer's scaled sigmoid function becomes too steep too quickly. Most likely, the issue involves multiple transformations interacting in a way that is difficult to predict. Broadening our hyperparameter search to include such variables would be one possible solution, though it is not clear how effective this brute force approach would be. We plan to systematically investigate this issue further in future work.

One limitation of our overall approach to explainability is that it relies on carefully engineered interpretable features. We were able to rely on those previously crafted by gaming-the-system experts using cognitive task analysis \cite{paquetteunderstandingexpertcoding2014}, but in many real-world scenarios, such features may not be available.

Another, less insurmountable, limitation of the current model is that the convolutional filters only allow for a fixed number of action steps. That is, the input may be of any length, but the patterns can only have at most three student action steps. This is an inherent limitation of the architecture, since the tensor containing the convolutional weights forces a fixed kernel size. One possible solution to explore in future work would be to provide separate but parallel convolutional layers, one for each desired sequence length. The feature maps, or output of each layer, could then be concatenated and passed through the thresholding layer as normal.

As we have mentioned, future work should seek to evaluate the explainability of the explanations derived from our convolutional filters. Following \cite{doshi-velezrigorousscienceinterpretable2017}, we plan to conduct two human-grounded evaluation experiments: forward simulation and counterfactual simulation. By asking participants to predict the model's output and to identify changes to inputs that will lead to different outputs, we can measure the extent to which our explanations are intelligible and faithful.

Finally, we plan to conduct an ablation study to measure the impact of each constraint on the model's accuracy and on the filters' interpretability. This will help us better understand the interplay between constraints and between interpretability and accuracy. It may also provide insights into the issue of gradients no longer flowing.

\section{Conclusion}

We have described a novel approach to creating a neural-network-based behavior detection model that is interpretable by design. By constraining the model's learning process through a series of regularization terms and architectural changes, we were able to create a model that learned patterns indicative of gaming-the-system behavior. These patterns emulate the structure of those identified by human experts, indirectly indicating that their interpretations are sound.

Importantly, the parameters pertaining to these patterns encompass all the learnable weights in the final model, providing 100\% explanatory potential to the model's inner workings. We demonstrated some possible ways in which these patterns can be used to create explanations for different potential audiences, all while remaining fully interpretable.

We have not yet conducted a systematic evaluation of these explanations, but we believe they have the potential to be both faithful and intelligible to human end-users. We indicated some promising ways to evaluate these claims via forward simulation and counterfactual simulation tasks as human-grounded evaluation experiments. We will conduct this evaluation in future work, along with an investigation into the model's gradient issues and an ablation study to better understand the interplay between flexibility and interpretability.

\section{Acknowledgments}

This study is supported by the National Science Foundation under Award \#1942962. Any conclusions expressed in this material do not necessarily reflect the views of the NSF.


\bibliographystyle{abbrv}
\bibliography{sigproc}

\balancecolumns
\end{document}